\documentclass[sigconf]{acmart}

\usepackage{listings}
\usepackage{xcolor}
\usepackage{makecell}
\usepackage{float}
\usepackage{tcolorbox}
\usepackage{enumitem}

\AtBeginDocument{%
  }

\copyrightyear{2025}
\acmYear{2025}
\setcopyright{acmlicensed}\acmConference[CIKM '25]{Proceedings of the 34th ACM International Conference on Information and Knowledge Management}{November 10--14, 2025}{Seoul, Republic of Korea}
\acmBooktitle{Proceedings of the 34th ACM International Conference on Information and Knowledge Management (CIKM '25), November 10--14, 2025, Seoul, Republic of Korea}
\acmDOI{10.1145/3746252.3761577}
\acmISBN{979-8-4007-2040-6/2025/11}

\usepackage[linesnumbered,ruled,vlined]{algorithm2e}

\begin{document}

\title{BiListing: Modality Alignment for Listings}

\author{Guillaume Guy}
\email{guillaume@airbnb.com}
\affiliation{%
  \institution{Airbnb}
  \city{San Francisco}
  \state{CA}
  \country{USA}
}

\author{Mihajlo Grbovic}
\email{mihajlo.grbovic@airbnb.com}
\affiliation{%
  \institution{Airbnb}
  \city{San Francisco}
  \state{CA}
  \country{USA}
}

\author{Chun How Tan}
\email{chunhow.tan@airbnb.com}
\affiliation{%
  \institution{Airbnb}
  \city{San Francisco}
  \state{CA}
  \country{USA}
}

\author{Han Zhao}
\email{han.zhao@airbnb.com}
\affiliation{%
  \institution{Airbnb}
  \city{San Francisco}
  \state{CA}
  \country{USA}
}

\renewcommand{\shortauthors}{Guillaume Guy, Mihajlo Grbovic, Chun How Tan, and Han Zhao.}

\begin{teaserfigure}
  \includegraphics[width=\textwidth, height=5.5cm, keepaspectratio]{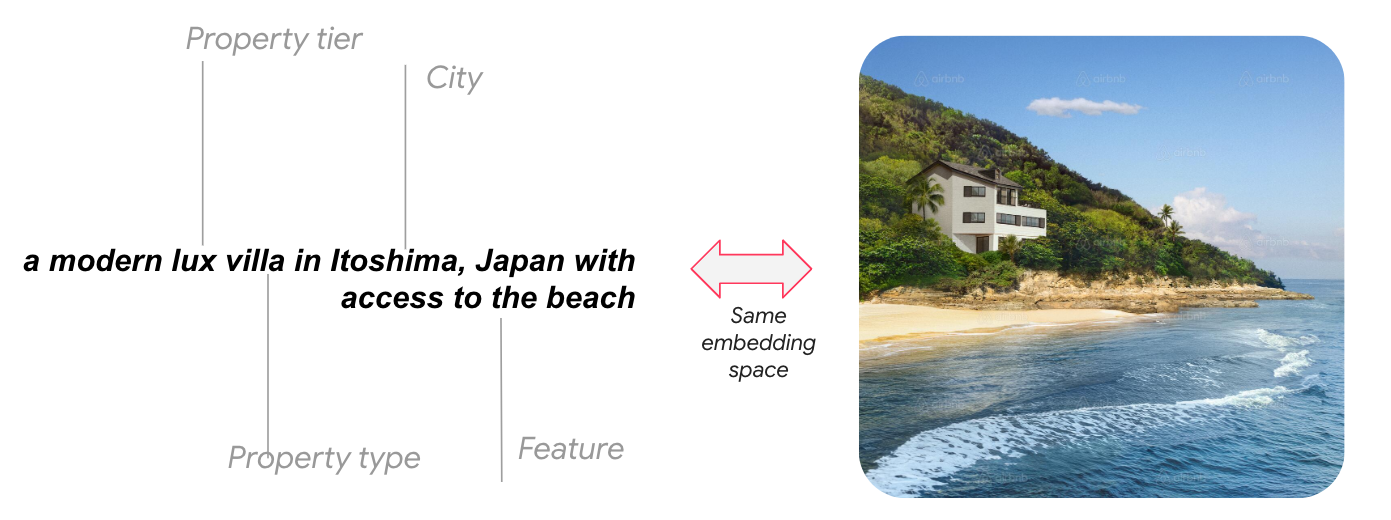}
  \caption{BiListing embeddings alignment}
  \label{fig:teaser}
  \Description{A representation of aligning listing photos and text description}
\end{teaserfigure}

\begin{abstract}
Airbnb is a leader in offering travel accommodations. Airbnb has historically relied on structured data to understand, rank, and recommend listings to guests due to the limited capabilities and associated complexity arising from extracting meaningful information from text and images. With the rise of representation learning, leveraging rich information from text and photos has become easier. A popular approach has been to create embeddings for text documents and images to enable use cases of computing similarities between listings or using embeddings as features in an ML model.

However, an Airbnb listing has diverse unstructured data: multiple images, various unstructured text documents such as title, description, and reviews, making this approach challenging. Specifically, it is a non-trivial task to combine multiple embeddings of different pieces of information, i.e. each image, each review, etc., to reach a single meaningful listing representation, especially if some of the embeddings lie in different spaces. Faced with such a problem, practitioners often resort to unprincipled approaches of averaging embeddings to produce a single one. However, this often results in an inaccurate representation due to loss of information in the averaging process.

This paper proposes BiListing, for Bimodal Listing, an approach to align text and photos of a listing by leveraging large-language models and pretrained language-image models. The BiListing approach has several favorable characteristics: capturing unstructured data into a single embedding vector per listing and modality, enabling zero-shot capability to search inventory efficiently in user-friendly semantics, overcoming the cold start problem, and enabling listing-to-listing search along a single modality, or both. 

We conducted offline and online tests to leverage the BiListing embeddings in the Airbnb search ranking model, and successfully deployed it in production, achieved 0.425\% of NDCB gain, and drove tens of millions in incremental revenue.

\end{abstract}



\maketitle

\section{Introduction}
\label{section:introduction}

In this section we introduce the problem, historical challenges and limitations of the existing solutions and our main contribution. In section \ref{section:related_works}, we describe related work in the area of item embeddings as well as multimodal representations. Section \ref{section:approach} outlines our approach in detail. In section \ref{section:experimental_results} we present the results, both quantitatively and qualitatively. Finally, in section \ref{section:application}, we describe practical considerations for successfully applying BiListing embeddings as features in the Airbnb search ranking model and demonstrate how multimodal embeddings can significantly improve guest search experience backed by a strong lift in bookings on Airbnb .

\subsection{Historical Context}

To generate listing representations, Airbnb has historically relied on user logs, i.e. co-clicked and co-wishlisted listings leading to bookings \cite{Grbovic18}, and less on unstructured data. However, user research shows that a guest spends a third of their time inspecting the images alone. The main challenge is to combine all the diverse listing content, i.e. images of different aspects and spaces of a home, title, description, and reviews, into a single representation.

For example, a listing may have 60 images, some of the backyard with lush garden and a pool, and some of the living room with Scandinavian furniture. Now consider that it also has a title and description which talk about the location, how to get there and finally reviews that talk about guest experiences. These are all very different aspects of the same listing that need to be meaningfully combined into a single representation. Averaging embeddings is inadequate because some aspects are more dominant than others. 

Recent advancements in contrastive learning allowed us to think about the problem as an alignment one, but there were still many practical challenges that needed to be solved which led to the work in this paper.

\subsection{CLIP and its limitations}
\begin{figure}[h]
  \centering
  \includegraphics[width=\linewidth]{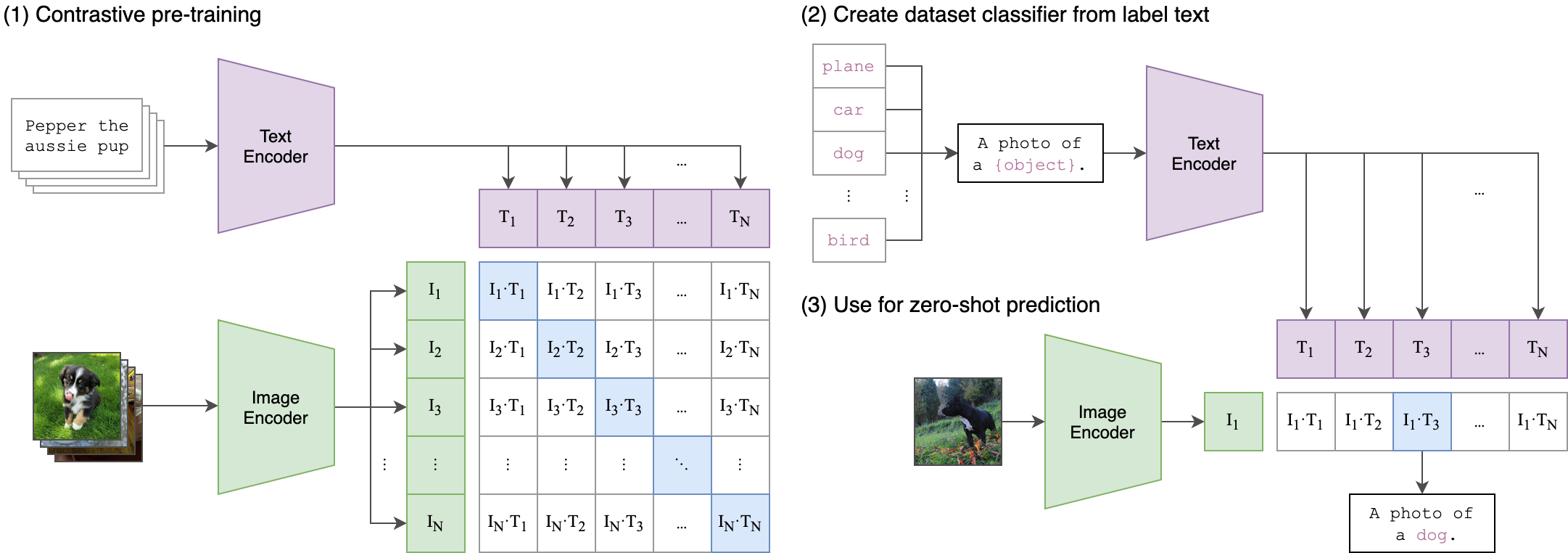}
  \caption{Contrastive learning diagram \cite{radford21}}
    \label{fig:contrastive_learning}
  \Description{The result of a free-text query using CLIP-based retrieval}
\end{figure}

\begin{figure}[h]
  \centering
  \includegraphics[width=\linewidth]{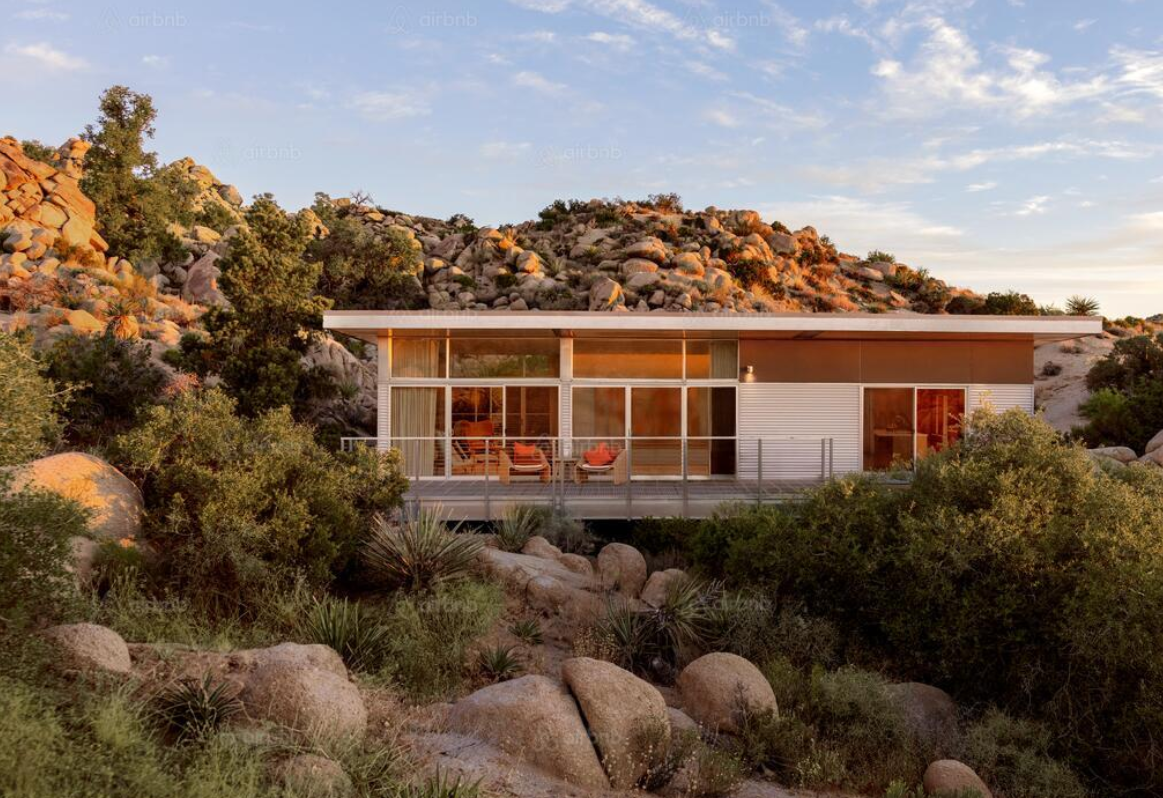}
  \caption{Illustration of an image result using from query: "A stunning  steel-and-glass home in the Yucca Valley"}
  \label{fig:yucca_exterior}
  \Description{The result of a free-text query using CLIP-based retrieval}
\end{figure}

With large-scale pretraining using image-text pairs \cite{radford21} (see Figure \ref{fig:contrastive_learning}) CLIP demonstrated that with orders of magnitude more web-sourced data \cite{schuhmann22}, it can provide semantic understanding of images, leading to a wide array of use cases such as image search using free text queries (see Figure \ref{fig:yucca_exterior}).

CLIP is trained using two encoders, one for captions and one for images in a 1:1 relationship. The output of both are then projected into a joint latent space, with the objective of bringing entities belonging to the same pair closer in the latent space.


Adapting CLIP to Airbnb's listings has several challenges:
\begin{itemize}[noitemsep, topsep=0pt]
\item CLIP only applies to a single photo:text pair whereas a listing has multiple photos and pieces of text. Therefore, free text listing searches such as "Home in the woods with a hot tub, sauna, and large screen TV" that cover multiple listing aspects that are potentially featured in different photos can lead to poor results. 
\item CLIP text encoder context is only 77 due to the inherent nature of captions which are usually single sentences, with front-loaded content. Listing descriptions alone are much longer and less front-loaded.
\item Pretrained text encoders are semantically tied to photo captioning with formats such as “a photo of \{\}” (e.g. “a photo of a dog”) which is an unnatural formulation for listing search application where queries would be more phrased like “A 2 bedroom villa in \{\} featuring \{\}” 
\end{itemize}

\subsection{Our contribution}
In this paper, we propose BiListings, an approach to align text and photos of a listing by leveraging large-language models and pretrained language-image models. Specifically, we propose adapting the Contrastive Learning Framework \cite{radford21} or CLIP to incorporate multiple aspects of the same listing, i.e. images of different parts of the listing and text attributes such as title, description and reviews that capture different aspects of the listing.

This paper is the first of its kind to expand the image-pair paradigm to more complex multimodal entities, and does so in a way that is easily reproducible by other practitioners in related applications where multiple aspects of the same item, that are possibly of different nature, need to be combined into a single item representation.

Furthermore, we present detailed steps to implement and reproduce our approach. We believe that intuitive and easily reproducible publications are important given the reproducibility crisis with some studies claiming that only a fraction of Machine Learning papers were reproducible\cite{albertoni2023reproducibility}

\section{Related Work}
\label{section:related_works}

OpenAI formulated the original CLIP problem in 2021 \cite{radford21}. Since, multiple improvements have been made. Zhang et al. \cite{zhang24} proposed another alignment process with coarse components to enhance the length of the CLIP context, but with added complexity.

In 2022, Flamingo \cite{Alayrac22} introduced a multimodal model capable of processing interleaved vision and text inputs for tasks such as visual question answering. However, its architecture poses three challenges for industrial-scale listing representation: (1) it is optimized for interleaved vision-language tasks rather than per-entity embedding use cases like retrieval and ranking; (2) it processes only a limited number of images—typically 5 during training—whereas listings on our platform may contain 64+ images; and (3) it relies on a large decoder-style architecture, making it more resource-intensive than BERT-style encoders, making Flamingo unsuitable for our latency- and throughput-constrained production use case. Similarly, BLIP \cite{Li22} only focuses on single image entities.

Video retrieval provides another interesting perspective, as videos are made up of frames, similarly to listings. Portillo-Quintero et al. \cite{portilloquintero21} experimented with mean-pooling of CLIP embeddings across video frames, showing strong video retrieval performance. Luo et al. \cite{luo21} went further and introduced a self-attention mechanism across video frames to learn temporal dependency, showing improved performance. 

Furthermore, Faysse et al. \cite{faysse24} proposed to see a document as a collection of patches, from which we can extract embeddings. While enabling multimodal representation, this approach still suffers from not having a unified embedding representation.

Finally, our work addresses the limitations of the approach presented in Grbovic et al. \cite{Grbovic18}. In that work the authors leverage user action sequence, such as clicks, wishlists, and bookings, to train high-quality listing embeddings. However, the approach suffers from a cold-start problem for listings that do not appear in many sessions. Moreover, the resulting embeddings can only be used to calculate similarities to other embeddings and for recommendation applications and are not searchable with natural language queries. In our work, we present an approach that is effectively able to use a large share of the listing content information from different modalities to create an embedding that is then both searchable with natural language queries and also usable for similarity-based recommendation applications. 

\section{Approach}
\label{section:approach}
\subsection{Objective}

We propose to formulate the Listing understanding as a modality alignment problem. As hosts upload their listing on Airbnb, the text, as well as the structured data they input provide details of the listings. See example in Table \ref{table:bilisting_yucca}

\begin{table}[h!]
    \centering
    \begin{tabular}{|c|c|}
     \hline
     \multicolumn{2}{|c|}{Inputs} \\
    \hline
     \textbf{Textual Inputs} & \textbf{Visual Inputs} \\
    \hline
        \begin{tabular}[c]{p{4cm}}
        \textbf{Title}: The Graham Residence on 20 Acres \\
        \textbf{Description}: Recently remodeled with gorgeous limestone floors and tiled the bathroom. The 1250-square-foot steel-and-glass home features two bedrooms and one bathroom. A generous living room has ample seating options and stunning views out of multiple sliding glass doors. Outside are two large decks with comfortable furniture and an outdoor dining table. A Jacuzzi-brand hot tub sits away from the house on a rock outcrop offering views in all directions. Other outdoor amenities include a propane [...]
        \end{tabular} &  
        \begin{tabular}{c}
          \includegraphics[width=0.4\linewidth]{figures/yucca_exterior.png} \\
          \includegraphics[width=0.4\linewidth]{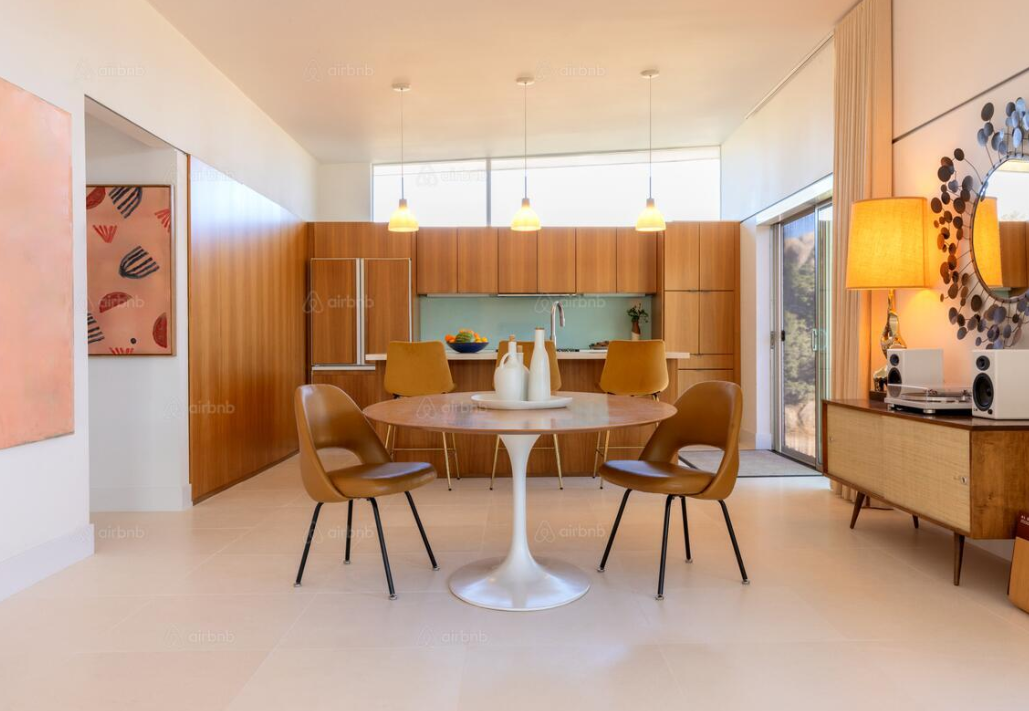} \\
          \includegraphics[width=0.4\linewidth]{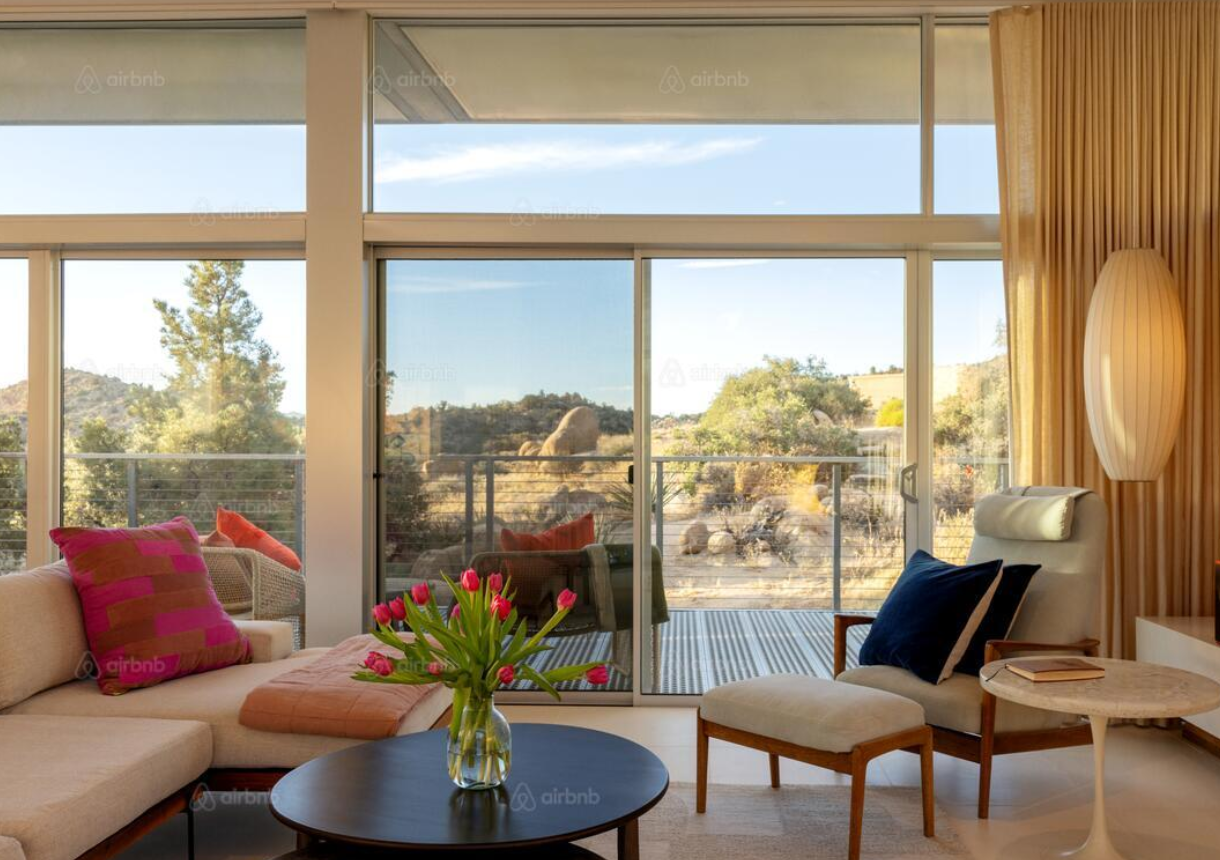} \\
        \end{tabular} \\
    \hline
    \end{tabular}
    \caption{side-by-side comparison of the multimodal inputs on a listing}
    \label{table:bilisting_yucca}
\end{table}

The text and photos are not randomly associated. In fact, hosts strive to give the salient details of their listings, featuring amenities, views, location, and vibe that transpire from the photos. In this example, the host highlights the recent renovation, with the materials (limestone floors and tiled bathroom), the sliding windows, and the expectations of quiet and wide surroundings (20 acres).

With this insight, we can take advantage of Airbnb’s large database of listings as training data without additional labels and adapt the CLIP image-text framework to train multimodal embeddings of a listing. 

\subsection{Architecture}
\begin{figure}[h!]
  \centering
  \includegraphics[width=\linewidth]{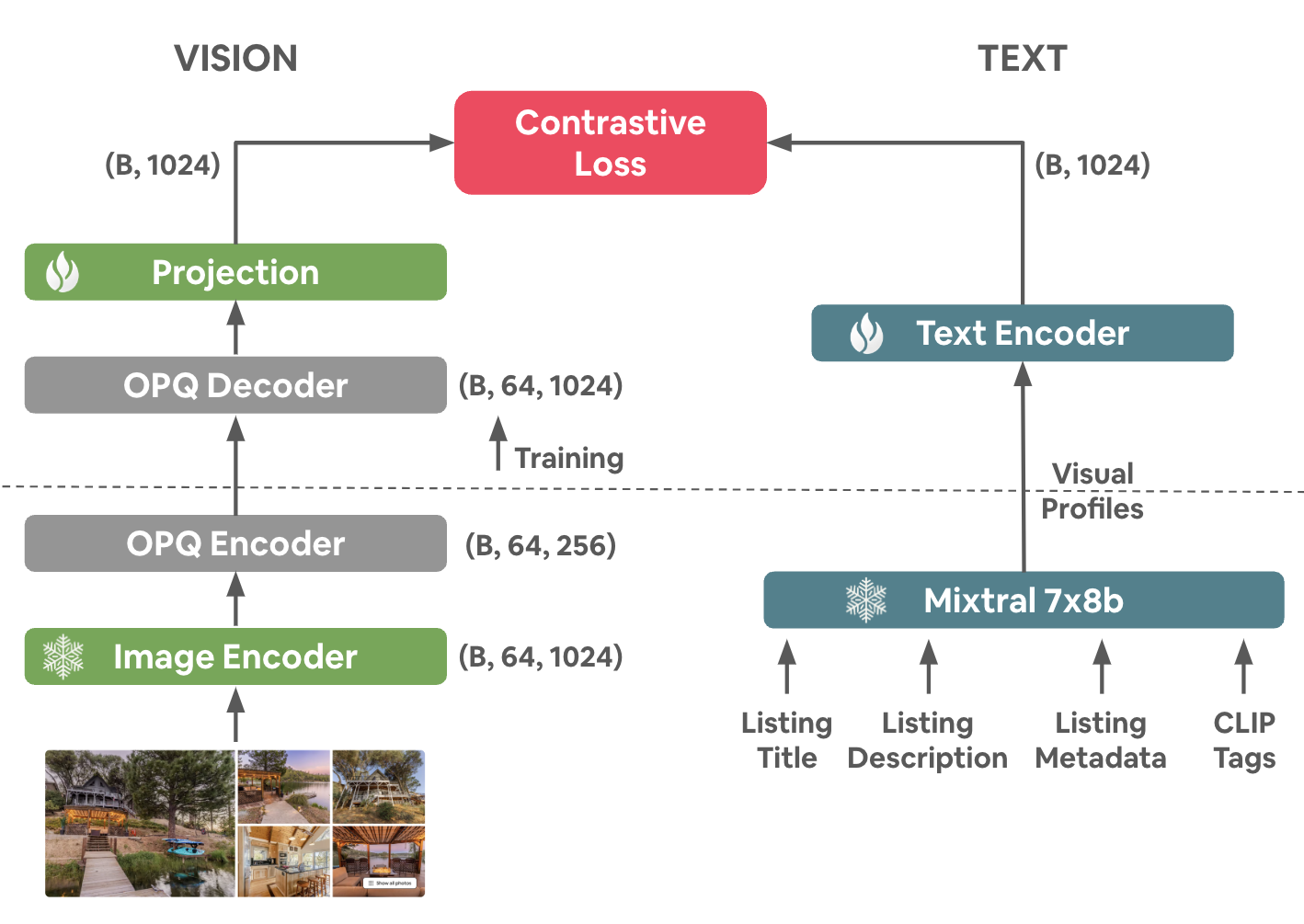}
  \caption{BiListing architecture using CLIP-style 2 tower architecture with vision (left) and text (right)
}
  \Description{2 tower architecture to align modalities}
  \label{figure:two-tower-architecture}
\end{figure}

We propose to reuse the 2 CLIP text-vision encoder tower approach in our BiListing model, as depicted in Figure \ref{figure:two-tower-architecture}. Compared to the standard CLIP model, we added three novel components to enable it to work at the listing level:
\begin{itemize}[noitemsep, topsep=0pt]
    \item In the vision tower, a trainable projection layer, using a Transformer architecture named "PhotoSet transformer", that consolidates multiple photo embeddings (up to 64\footnote{We chose 64 as 98 percent of Airbnb's listings have 64 or less photos}) into a single projection
    \item An aggressive embedding compression logic using Optimal Product Quantization (OPQ) to optimize bandwidth consumption 
    \item The generation of Visual Profiles, which summarize the listing's text description grounded with CV indicators into a short, dense, visual description of the listing
\end{itemize}

Zooming into the vision tower, our experiments show that a transformer encoder with positional encoding performs best. With this architecture, a dense embedding of an image can be interpreted as a token and all the photos make up a sentence. Within such a comparison, the EOT token is simply the embedding output of the last photo. In practice, positional encoding captures the fact that hosts tend to add the most important photos first, and photo of the surrounding areas last.

The pseudo-code associated with the proposed architecture is formulated in Algorithm \ref{alg:alignment_algorithm}.
 
\SetKwComment{Comment}{/* }{ */}
\begin{algorithm}
\KwIn{
\begin{itemize}[noitemsep, topsep=0pt]
    \item CLIP text encoder $T_{enc}$
    \item CLIP photo encoder $P_{enc}$
    \item PhotoSet encoder $PS_{enc}$
    \item Embedding dimensions (e.g., 1024) $d$
    \item Number of photos per listing $P$
    \item Batch size $B$
    \item Dataset $D$
\end{itemize}
}
\KwOut{$PS_{enc}$, $T_{enc}$} 
Initialize the CLIP encoders $T_{enc}$, $P_{enc}$ from the pretrained checkpoints; 
Freeze $I_{enc}$ (and $T_{enc}$ if Coarse Alignment)\;
Initialize the PhotoSet encoder $PS_{enc}$ weights with normal distribution\;
\ForEach{$(photos, visual\_prof) \in D$}{
    $photo\_ftrs \gets$ $P_{enc}$.encode($photos$) \Comment*[r]{[$B$,$P$,$d$]} 
    $PS\_ftrs \gets$ $PS_{enc}$.project($photo\_ftrs$)\Comment*[r]{[$B$,$d$]} 
    $text\_ftrs \gets$ $T_{enc}$.encode($visual\_prof$)\Comment*[r]{[$B$,$d$]} 
    $logits \gets PS\_ftrs \cdot text\_ftrs.T$ \Comment*[r]{[$B$,$B$]} 

    $labels = np.arange(B)$ \Comment*[r]{[$B$]}
    $loss\_p = cross\_entropy(logits, labels, axis=0)$\;
    $loss\_t = cross\_entropy(logits, labels, axis=1)$\;
    $loss = \frac{loss\_p + loss\_t}{2}$\;

    $T_{enc}$.backward()\;
    $PS_{enc}$.backward()\;
    
}
\Return $T_{enc}$, $PS_{enc}$ \;
\caption{Algorithm to align of Listings modalities}
\label{alg:alignment_algorithm}
\end{algorithm}

\subsection{Visual Text Profiles}

Airbnb collects a large number of disparate textual information (see \ref{table:visual_profile}): listing title, description, location but also Airbnb-generated signals (e.g. CLIP implicit signals, detected amenities) and structured data which all need to be aggregated into single, semantically correct, and natural prompt. Moreover, the prompt must contain only visual elements to facilitate the alignment with photos. For instance, Air Conditioning, or nearby POIs, are not visible features and should be excluded. 

To this end, we used a Large Language Model to generate summarized, dense text descriptions of listings capturing the visual salient points for guests, using the words of the hosts. We denote the output "visual profiles", as they represent a consolidated summary of the visual aspect of the listings.

Specifically, we used Mixtral 8x7b \cite{jiang24} in a zero-shot setting with a step-by-step recipe prompt for its open-weight availability and good balance inference cost vs. quality. We found that prefilling the response with ''A N bedroom, M bathroom'' produces the best outcome.

\begin{table}[h!]
\centering
\begin{tabular}{|p{2cm} |p{5cm}|}
\hline
\textbf{Source} & \textbf{Description} \\
\hline
\textbf{Title \& Description} & Host-provided title and description. \\
\hline
\textbf{Location} & Listing location as Airbnb 's searches are very local \\
\hline
\textbf{CLIP implicit signals} & 
Visual tags using implicit feedback collected from internal users. 
\textbf{Example:} ``A picture of a tiny house along the river'' \\
\hline
\textbf{Visually Detected Amenities} & 100+ amenities visually through object detection for small amenities (e.g., BBQ) and image classification (e.g., golf simulator). \\
\hline
\textbf{Visual rating} & A model generated rating indicating the visual appeal of the listing. \\
\hline
\textbf{Structured data} & Listing Category \\
\hline
\end{tabular}
\caption{Inventory of signals going into a visual profile}
\label{table:visual_profile}
\end{table}

As an illustration, the following output was generated for the listing from Table \ref{table:bilisting_yucca}:
''A stunning 2 bedroom 1250-square-foot recently-remodeled steel-and-glass home with gorgeous limestone floors and tiled bathroom featuring a generous living room with ample seating options and stunning views out of multiple sliding glass doors, a jacuzzi, balcony, and a propane fire pit. Located on 20 acres of desert landscape in Yucca Valley, California, USA, the property is nestled among weathered boulders, ancient junipers, and desert oaks.''

\subsection{Quantization}

To accommodate multiple images per listings and still maintain a large batch size as proven empirically necessary \cite{zhai23}, normalized image embeddings are encoded offline through an Optimal Product Quantization (OPQ) technique \cite{Ge13} and decoded as part of the training preprocessing.

Specifically, we use OPQ with settings of 256 centroids and a rotation matrix of output dimension of 1280. Each embedding is therefore reduced to 256 uint8, a 97\% reduction in size at a loss of 9.9\% of L2 mean error. Our experience shows that this loss is acceptable and keeps high fidelity to the original representation.

\section{Experimental Results}
\label{section:experimental_results}

In this section, we provide a detailed description of BiListing training procedure, offline evaluations we conducted, a summary of the results, and post training steps that were needed to adapt BiListing embeddings for application usages in listing search and ranking. 

For easier reproduction, we described details of each steps we took, practical challenges we faced, and how we overcame them. 

A dedicated section on BiListing embeddings path to production in listing search and ranking as features of the ranking ML model will follow afterwards.

\subsection{Data Preparation}

As noted in previous research \cite{datacomp23, fang23}, data quality plays a significant role in modeling performance. We pay special attention to the data quality of the text-based visual profiles and apply multiple data processing and validation steps to our training and evaluation pipeline. The training and evaluation datasets exclude:
\begin{itemize}
\item  Listings with less than 5 photos
\item  Listings with short visual profiles (<50 characters)
\item  A preliminary model trained on a large listing corpus is used to filter out low alignment listings. We use 0.3 as a threshold. 
\end{itemize}

Finally, the raw visual text profile LLM outputs are lightly edited using predefined regular expressions. For instance, the number of bathrooms is removed, and "0 bedroom" is edited to "studio".

Together, the rules result in a reduction in 35\% in the size of the dataset, now in the low tens of millions of listings. 

\subsection{Training Details}

 For the training, we extract the text encoder from a pretrained CLIP checkpoint (laion/CLIP-ViT-H-14-laion2B-s32B-b79K) and finetune it using the open\_clip library \cite{ilharco21} on 8XA100 for 15 hours, an accumulation size of 8, a weight decay of 0.2, a warmup of 10,000 steps and an individual batch size of 512 for a global batch size of 32k. The training uses Adam \cite{kingma17} coupled with a cosine scheduler. The model is trained using Ray TorchTrainer on top of the open\_clip library.

The Photoset Transformer hyperparameters is tuned with the results below. The best model had 4 layers and 4 attention heads, and underwent a two-step training process:\begin{itemize}
    \item \textbf{Coarse Alignment}: Frozen text encoder for 140 epochs with a learning rate of $1e-4$. The main purpose of this step is to pretrain the PhotoSet encoder aligning on the native CLIP representation
    \item \textbf{Fine Alignment}: The next 80 epochs had a learning rate of $2e-5$ with the last 5 layers of the text encoder being unfrozen. The LayerNorm running stats in text tower for any locked layers were frozen.
\end{itemize}

\subsection{Evaluation Results}

\subsubsection{Model Evaluation}

We set aside a random 1\% of all listings as a holdout and evaluate performance metrics on this holdout. We evaluated the following metrics and present the results in Table \ref{tab:experiment_results} and Table \ref{tab:experiment_results_recall}: \begin{itemize}
    \item \textbf{Mean rank (text to image) (MR T->I),} the average rank position of the correct match in the retrieval set
    \item \textbf{Text to image recall (@ 1,5, 10),} the proportion of correct matches ranked in the top-K.
    \item \textbf{Image to text recall (@ 1,5, 10)}
\end{itemize}

\begin{table}[h!]
\centering
\renewcommand{\arraystretch}{1.5}
\setlength{\tabcolsep}{8pt}
\begin{tabular}{|c|c|c|}
\hline
\textbf{Experiment} & \textbf{Loss} & \textbf{MR T->I (best)}  \\
\hline
\multicolumn{3}{|c|}{\textbf{Baseline}} \\
\hline
MLP (baseline) & InfoNCE & 26.5 \\
\hline
\hline
\multicolumn{3}{|c|}{\textbf{PhotoSet Transformer}} \\
\hline
\multicolumn{3}{|c|}{\textit{without coarse alignment}} \\
\hline
PS Trans (4 layers, 8 heads) & InfoNCE & 20.2 \\
\hline
PS Trans (8 layers, 8 heads) & InfoNCE & 20.5 \\
\hline
PS Trans (4 layers, 8 heads) & SigLIP \cite{zhai23} & 8.43 \\
\hline
\multicolumn{3}{|c|}{\textit{with coarse alignment}} \\
\hline
\textbf{PS Trans (4 layers, 4 heads)} & \textbf{SigLIP} & \textbf{6.17} \\
\hline
\end{tabular}
\caption{Experiment results of different model architecture}
\label{tab:experiment_results}
\end{table}

The evaluation results reveal several useful improvements, in descending order of impact:\begin{itemize}
    \item Switching over to SigLIP as a loss function
    \item Introducing a coarse alignment step followed by a fine tuning step with lower learning rate
    \item Replacing the MLP baseline with the Photoset transformer
\end{itemize}

For the best model, we provide more metrics in Table \ref{tab:experiment_results_recall}.
\begin{table}[h!]
\centering
\renewcommand{\arraystretch}{1.5}
\setlength{\tabcolsep}{8pt}
\begin{tabular}{|c|c|}
\hline
\textbf{Metric} & \textbf{Value} \\
\hline
Mean rank, (text-> image) & 6.2 \\
Recall @ 1 (text-> image) & 58\% \\
Recall @ 5 (text-> image) & 85\% \\
Recall @ 10 (text-> image) & 91\% \\
Mean rank, (image -> text) & 6.0 \\
\hline
\end{tabular}
\caption{Experiment results of the best model setting on a 10k randomly selected listing sample from the holdout}
\label{tab:experiment_results_recall}
\end{table}

\subsubsection{Model Comparison}

We perform a performance comparison against the in-production embeddings ("Legacy"), an updated version of Grbovic et al. \cite{Grbovic18} original work. To perform the comparison, we choose a wide range listing structured fields as targets and evaluate using k-nn probes trained on the training set. We also provide a zoom-in onto a sample of Airbnb Categories, Airbnb's structured property offering.

\begin{table}[h]
\centering
\begin{tabular}{lcc}
\toprule
Attribute & BiListing (vision) & Legacy \\
\midrule
Urban/Rural Density       & \textbf{0.64} & 0.41 \\
Space Type (apt, ...)     & \textbf{0.62} & 0.46 \\
Capacity (\# people)      & \textbf{0.57} & 0.46 \\
Bedroom count             & \textbf{0.55} & 0.49 \\
Bathroom count            & \textbf{0.75} & 0.72 \\
ADR percentile            & \textbf{0.29} & 0.27  \\
Pro Host (Y/N)            & 0.655 & \textbf{0.74} \\
Listing Tenure (days)     & 0.491 & \textbf{0.62}  \\
Host is superhost         & 0.53 & \textbf{0.68} \\
Review count (bucketized) & 0.37 & \textbf{0.64}  \\
Bookings count (bucketized) & 0.26 & \textbf{0.56}  \\
\bottomrule
\end{tabular}
\caption{Listing structured attributes accuracy comparison}
\end{table}
\vspace{-2em}

\begin{table}[h]
\centering
\begin{tabular}{lcc}
\toprule
Attribute & BiListing (vision) & Legacy \\
\midrule
A-Frame     & \textbf{0.87} & 0.80 \\
Pool        & \textbf{0.88} & 0.81 \\
Cabin       & \textbf{0.86}  & 0.81 \\
Lakehouse   & \textbf{0.87} & 0.81 \\
Cave        & \textbf{0.87} & 0.83 \\
Tropical    & \textbf{0.89} & 0.85 \\
Beachfront  & \textbf{0.88 }& 0.85 \\
\bottomrule
\end{tabular}
\caption{Airbnb categories accuracy comparison}
\end{table}

BiListing performs favorably on most attributes but not all. Unsurprisingly, BiListing performs best for tasks which most benefit from vision (e.g. Urban vs. Rural).

\subsection{Ablation Study}
\label{subsection:ablation-study}

To understand the role played by the different components of this complex workflow, we conduct two ablation studies, one on input quality and the other on output quality:
\begin{itemize}
    \item \textbf{Quantization of CLIP embeddings}: The training procedure uses Optimal Product Quantization. We compare the approach against the more common Product Quantization and analyze the magnitude of the errors
    \item \textbf{Dimensionality Reduction of BiListing embeddings}: The integration into the Ranker uses a reduced number of dimensions . We present a study on how the number of dimensions affect the ranker's performance
\end{itemize}

\subsubsection{Quantization of CLIP embeddings}
\label{subsubsection:quantization-of-clip-embeddings}

We evaluated multiple quantization strategies for the vision tower CLIP embeddings: Product Quantization (PQ) \cite{Jégou11} and Optimal Product Quantization (OPQ) \cite{Ge13}. Table \ref{tab:experiment_results_dist_loss}, indicates that the random rotation introduced in OPQ achieves a significant improvement in the compression quality, with over 54\% error reduction at the P50.

\begin{table}[h!]
\centering
\setlength{\tabcolsep}{3pt}
\begin{tabular}{c | c | c | c | c | c | c}
\hline
 & $p_5$ & $p_{25}$ & $p_{50}$ & $p_{75}$  & $p_{90}$ & $p_{99}$ \\
\hline
\textbf{PQ} & 18.75 & 19.87 & 20.93 & 22.37 & 24.21 & 29.03 \\
\textbf{OPQ} & 8.30 & 8.89 & 9.49 & 10.36 & 11.64 & 16.46 \\
\textbf{L2 Error reduction (\%)} & 55.73 & 55.27 & 54.68 & 53.68 & 51.92 & 43.31 \\
\hline
\end{tabular}
\caption{Distribution of loss following OPQ vs. PQ encoding}
\label{tab:experiment_results_dist_loss}
\end{table}
\vspace{-2em}
\subsubsection{Dimensionality Reduction of BiListing embeddings}
\label{subsubsection:quantization-of-biListing-embeddings}
Each tower produces an embedding of dimension 1024, too large to use in the production search ranking model as a feature due to various infrastructure requirements, latency requirements and scalability. To address this, we used Principal Component Analysis \cite{Jolliffe02} followed by quantization.

\begin{figure}[h]
  \centering
  \includegraphics[width=\linewidth, trim=0 20 0 30, clip]{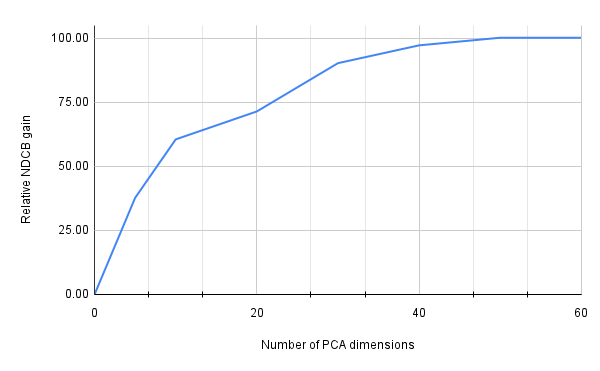}
  \caption{NDCB vs. top n PCA dimensions}
  \Description{A plot showing NDCB values as a function of the number of top PCA dimensions.}
  \label{figure:ndcb_gain_by_pca_dimension}
\end{figure}

For offline evaluation of the Airbnb's search ranking model, we use Normalized Discounted Cumulative Gain (NDCB) \cite{Jarvelin02} with binary relevance score: whether the listing was booked. Figure \ref{figure:ndcb_gain_by_pca_dimension} shows the relative offline NDCB gain, compared to the final model, when adding different dimensions of the BiListing embeddings as features. We ended up adding just the first 40 PCA dimensions to the production model as we observe diminishing returns with additional dimensions.

\vspace{-2em}

Finally, we further investigate whether we can further compress the BiListing embedding through scalar quantization. No statistically significant differences were observed using just 8-bits precision with the full 32-bits float precision. 

In conclusion, using the first 40 dimensions of 8-bit-encoded BiListing embeddings proves to be enough.

\section{Application}
\label{section:application}

\subsection{Integration into Listing Search Ranking}
\label{subsection:integration-into-ranking}

\begin{tcolorbox}[colback=gray!5!white,colframe=black!75!black,title=Deployment Highlights]
\begin{itemize}[left=0pt,labelsep=0.5em]
    \item \textbf{Production deployment:} BiListing provides the core listing representation of Airbnb's core search ranking system.
    \item \textbf{Latency-aware design:} Embeddings are quantized (PCA + byte quantization) to 40 bytes per listing, supporting real-time scoring.
    \item \textbf{Cold start remediation:} Since BiListing uses photos only, its design is more robust to the cold start problem.
    \item \textbf{Integration:} Fully integrated into feature pipelines with automated daily embedding refresh.
    \item \textbf{Business impact:} +0.425\% offline NDCG gain; A/B test shows statistically significant uplift in Uncancelled Bookers (p = 0.025), resulting in \textbf{tens of millions incremental revenue/year}.
\end{itemize}
\end{tcolorbox}

We integrated BiListing into the core search ranking model \cite{Tan23, Tang24} powering Airbnb’s results page, addressing two main challenges: embedding freshness and compactness. A daily pipeline recomputes CLIP embeddings for new photos, applies OPQ quantization, and updates BiListing vectors.

As described in section~\ref{section:approach}, each listing is encoded into 1024-dimensional vision and text embeddings trained in a shared latent space. For ranking, we use only the vision embedding—cheaper to generate and independent of text availability—arranged as a sequence capturing visitors’ short- and long-term interactions with Airbnb inventory.

Following our quantization ablation (\ref{subsubsection:quantization-of-biListing-embeddings}), we reduced embeddings from 1024 to 256 dimensions via PCA, observed diminishing returns beyond the 40th component (Figure~\ref{figure:ndcb_gain_by_pca_dimension}), and applied 8-bit scalar quantization. The final representation is just 40 bytes per listing.

This compact embedding delivered a +0.425\% offline NDCG gain—above the +0.3\% threshold \cite{Tan23} typically needed for statistically significant booking lift—and, in an online A/B test, increased Uncanceled Bookers (p = 0.025), yielding tens of millions in incremental revenue. By transitivity, these results also confirm additional gains over the listing embeddings in \cite{Grbovic18}.

\subsection{Internal Listing search}
\setlength{\parskip}{0pt}
\label{subsection:free-text-search-of-listing}

To help users find listings with certain attributes, we deploy an online application using Streamlit for the UI and FAISS \cite{douze24}, on Airbnb's Sandcastle platform \cite{Miller24}. All active listings have their visual profile (text) and CLIP embeddings (photos) encoded into two sets of embeddings. A third embedding, denoted multimodal is inferred by mean-pooling the first two. Airbnb internal employees can query listings two ways: 
\begin{itemize}[noitemsep, topsep=0pt]
    \item \textbf{Free text search}. For instance "A 2 bedroom apartment in La Rochelle with a view on the harbor"
    \item \textbf{Listing-to-Listing}. Using a listing embedding to perform a nearest-neighbor search. 
\end{itemize} 
For each query, the user can select which of the 3 modalities to use. We provide a representative sample of (top-1) results in Table \ref{tab:free-text-search} for a sample of free text queries:
\begin{table}[H]
\centering
\renewcommand{\arraystretch}{1.5}
\setlength{\tabcolsep}{7.5pt}
\begin{tabular}{|p{2.5cm}|p{5cm}|}
\hline
\centering \textbf{Query} & \makecell{\textbf{Image}} \\
\hline
\centering A stunning treehouse in Japan & \makecell{\includegraphics[width=0.6\linewidth]{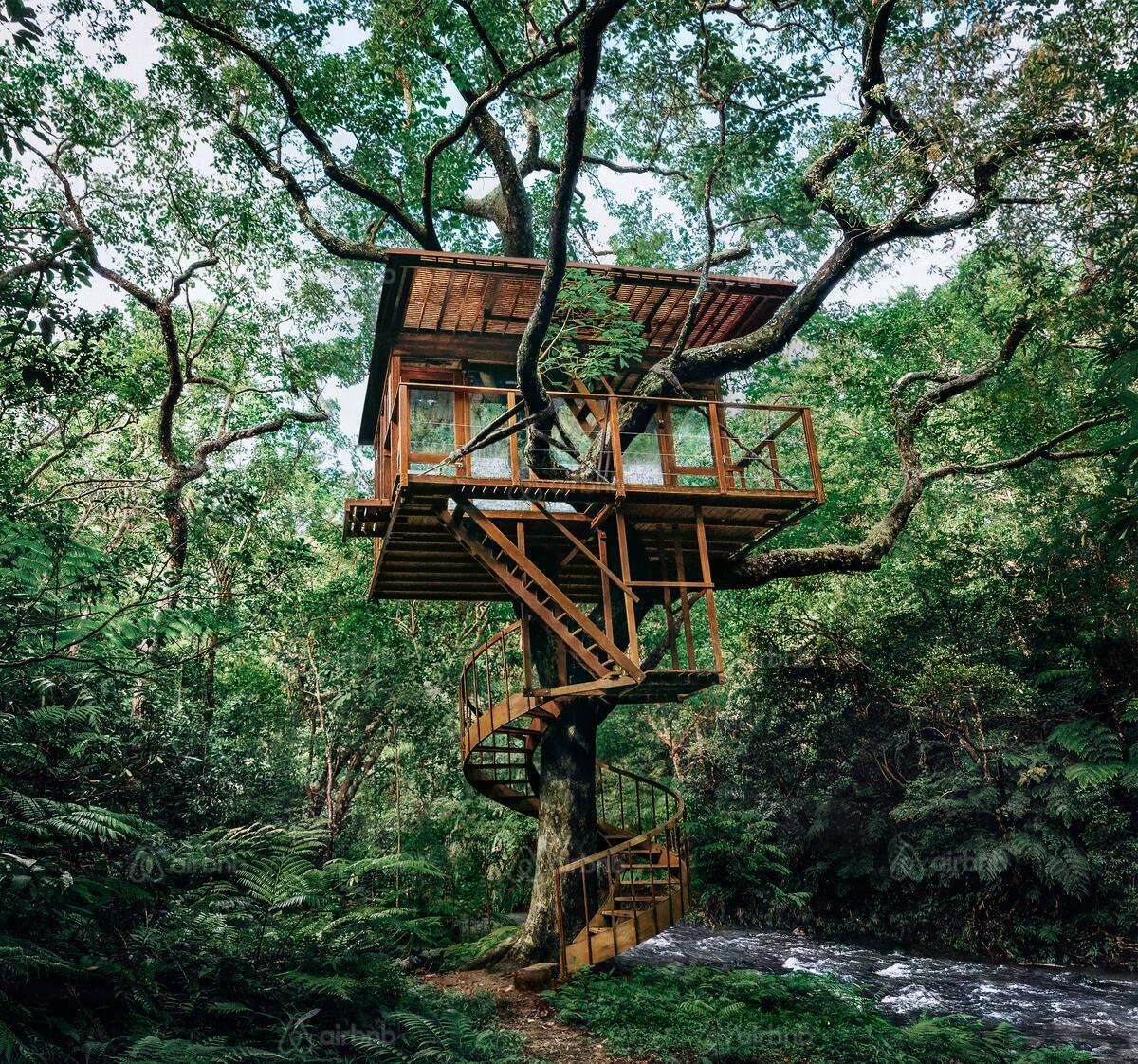} \\ \centering}  \\ 
\hline
\centering a visually appealing Ryokan in Japan &  \makecell{\includegraphics[width=0.6\linewidth]{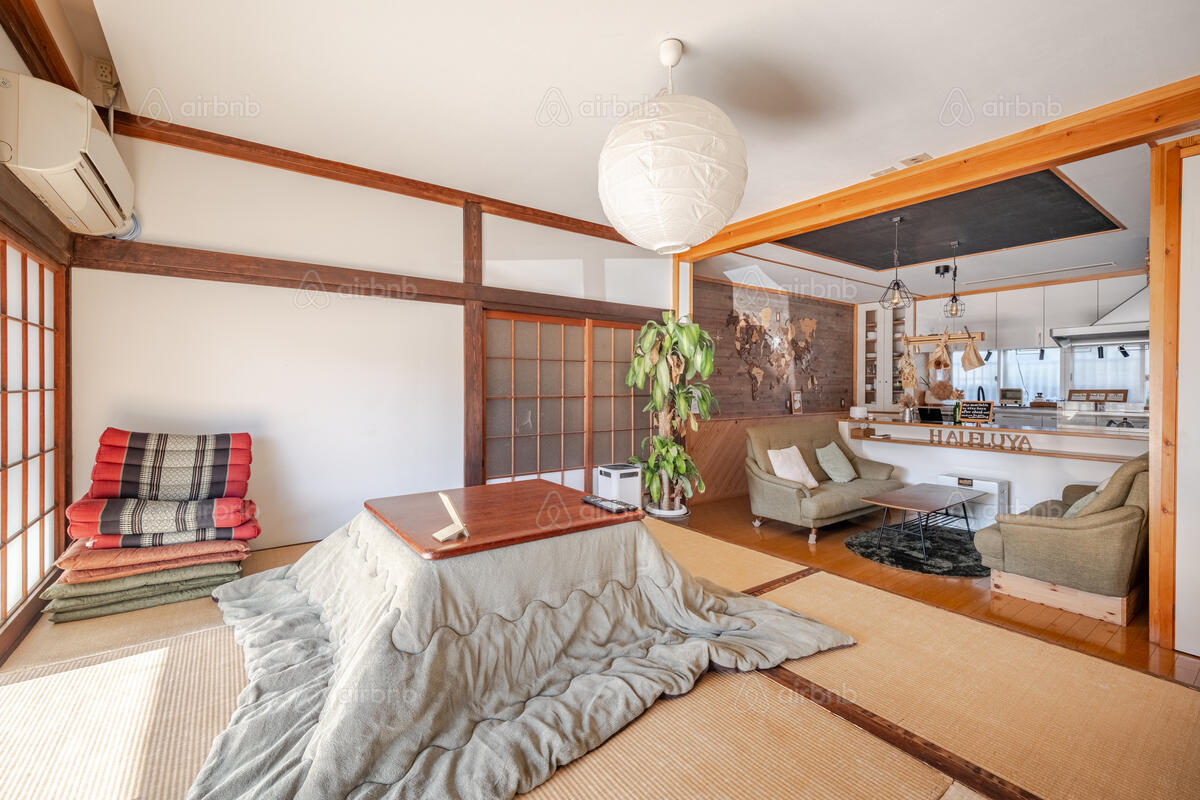} \\ \centering }\\ 
\hline
\end{tabular}
\caption{Free-text search results in Airbnb's inventory}
\label{tab:free-text-search}
\end{table}
\vspace{-3em}

Due to privacy concerns, we cannot provide listing-to-listing search result samples using a listing input query. However, the results show that the properties are indeed similar in many respects:
\begin{itemize}
    \item Location (e.g. the Caribbeans)
    \item Property Type (e.g Villa)
    \item top amenities (e.g. pool, view of the beach)
\end{itemize} 

\section{Conclusion}

We proposed BiListing embeddings, extending the standard modality alignment approach through contrastive learning to a more complex entity - Airbnb's listings. We demonstrate the usefulness of the BiListing embedding in understanding the Airbnb's listings better through two downstream applications: Airbnb's search ranking model and as a standalone free text search application for internal Airbnb employees. Our special emphasis was on explaining our training in detail such that it can be easily reproduced by other practitioners who we believe will find it useful for their applications.

We would also like to call out that the illustrative photos in this paper had to comply with Airbnb's privacy policy and had to be changed. We believe that this change does not change the character of the findings.

\section{Future Work}
While BiListing demonstrates measurable gains in production quality for complex multi-image entities such as Airbnb's listings, several avenues remain open for exploration.
First, evaluating how techniques for context extension such as Long-CLIP \cite{zhang24} or Tulip \cite{najdenkoska25} could further enhance the model.
Second, expanding ablation studies on the choice of large language model for Visual Text Profiles and the role of positional encoding in the PhotoSet Transformer would strengthen the understanding of design trade-offs.
Finally, deeper analysis of cold-start performance and integration strategies within the ranking system may yield additional performance gains in practical deployments.

\section{AI Usage Disclosure}
AI tools were used to assist with grammar checking and minor rephrasing. 
The authors remain fully responsible for the accuracy, originality, and integrity of all content presented in this work.

\bibliographystyle{ACM-Reference-Format}
\balance
\bibliography{citations}

\appendix

\end{document}